%% file: main.tex
\documentclass[a4paper,twoside]{article}

\usepackage{epsfig}
\usepackage{subcaption}
\usepackage{calc}
\usepackage{amssymb}
\usepackage{amstext}
\usepackage{amsmath}
\usepackage{amsthm}
\usepackage{multicol}
\usepackage{pslatex}
\usepackage{apalike}
\usepackage{algorithm2e}
\usepackage[bottom]{footmisc}
\usepackage{SCITEPRESS}     
\usepackage{booktabs}

\begin{document}

\title{From Data Lifting to Continuous Risk Estimation: 
A Process-Aware Pipeline for Predictive Monitoring of Clinical Pathways}

\author{\authorname{Pasquale Ardimento\sup{1}\orcidAuthor{0000-0001-6134-2993}, Mario Luca Bernardi\sup{2}\orcidAuthor{0000-0002-3223-7032} Marta Cimitile\sup{3}\orcidAuthor{0000-0003-2403-8313} and Samuele Latorre\sup{1}\orcidAuthor{0000-0000-0000-0000}}
\affiliation{\sup{1}Department of Informatics, University of Bari Aldo Moro, Bari, Italy}
\affiliation{\sup{2}Department of Engineering, University of Sannio, Benevento, Italy}
\affiliation{\sup{3}Department of Economy and Law, UnitelmaSapienza University, Rome, Italy}
\email{pasquale.ardimento@uniba.it, bernardi@unisannio.it, marta.cimitile@unitelmasapienza.it, s.latorre11@studenti.uniba.it}
}


\abstract{This paper presents a reproducible and process-aware pipeline for predictive monitoring of clinical pathways. The approach integrates data lifting, temporal reconstruction, event log construction, prefix-based representations, and predictive modeling to support continuous reasoning on partially observed patient trajectories, overcoming the limitations of traditional retrospective process mining. The framework is evaluated on COVID-19 clinical pathways using ICU admission as the prediction target, considering 4,479 patient cases and 46,804 prefixes. Predictive models are trained and evaluated using a case-level split, with 896 patients in the test set. Logistic Regression achieves the best performance (AUC 0.906, F1-score 0.835). A detailed prefix-based analysis shows that predictive performance improves progressively as new clinical events become available, with AUC increasing from 0.642 at early stages to 0.942 at later stages of the pathway. The results highlight two key findings: predictive signals emerge progressively along clinical pathways, and process-aware representations enable effective early risk estimation from evolving patient trajectories. Overall, the findings suggest that predictive monitoring in healthcare is best conceived as a continuous, dynamically aware process, in which risk estimates are progressively refined as the patient journey evolves.}

\onecolumn \maketitle \normalsize \setcounter{footnote}{0} \vfill

{\small
\noindent
This manuscript has been submitted to the 21st IEEE International Conference on Software Technologies (ICSOFT 2026). 
It has not undergone peer review, copy-editing, or typesetting. 
The final authenticated version, if accepted, will be available through the conference proceedings and associated DOI.
}

\section{\uppercase{Introduction}}
\label{sec:introduction}
Digital health aims to improve healthcare delivery through the integration of information technologies, data-driven services, monitoring systems, and decision-support mechanisms. The COVID-19 pandemic further emphasized the need for reliable digital health infrastructures capable of supporting healthcare professionals in dynamic, data-intensive, and high-risk scenarios. Hospitals routinely collect large volumes of heterogeneous clinical data from electronic health records and monitoring systems. Despite this, transforming such data into reliable decision-support indicators remains challenging due to incompleteness, temporal inconsistencies, and fragmentation. Addressing these issues requires systematic data transformation processes (data lifting) to enable downstream analysis and predictive modeling.

Process mining provides methods for reconstructing and analyzing real-world clinical pathways from event data. However, many healthcare process mining studies remain primarily retrospective: they analyze completed patient trajectories to discover process models, identify bottlenecks, or study pathway variability. While this is valuable, digital health systems increasingly require capabilities for continuous and prospective monitoring, where patient states are updated as the pathway evolves and decisions must be supported under partial and evolving information.

In this context, predictive monitoring represents a key extension of process mining, enabling the estimation of future outcomes from partially observed trajectories. However, existing approaches often focus on predictive models in isolation, while paying less attention to the integration of data transformation, process-aware representations, and reproducible experimental pipelines.

This paper proposes a reproducible and process-aware pipeline for predictive monitoring of clinical pathways. Rather than introducing a new predictive model, the focus is on the design of an end-to-end pipeline that integrates data lifting, temporal reconstruction, event log construction, prefix-based representations, and predictive modeling within a unified and traceable framework. This design enables continuous risk estimation from partially observed patient trajectories and supports the analysis of how predictive signals emerge over time. The main contributions of this paper are:
\begin{itemize}
\item A reproducible pipeline for transforming heterogeneous clinical records into temporally consistent event logs and process-aware representations suitable for predictive monitoring.
\item A prefix-based monitoring framework that enables continuous prediction from partially observed clinical pathways.
\item An empirical evaluation showing how predictive performance evolves along clinical pathways, providing insights into the progressive emergence of predictive signals and the role of process-aware representations.
\end{itemize}

\section{Related Work}
\label{sec:related}

The proposed work intersects three main research areas: process mining in healthcare, event log construction and data quality, and predictive process monitoring.

Process mining provides techniques for discovering, monitoring, and improving processes from event logs. Foundational work by van der Aalst established its role as a bridge between data science and process science~\cite{pm_aalst2016}. In healthcare, it has been applied to patient flow analysis, clinical pathway discovery, bottleneck identification, and conformance checking. Rojas et al.~\cite{pm_rojas2016} review its application in healthcare, while Muñoz-Gama et al.~\cite{pm_munoz2022} discuss key challenges such as data complexity, privacy, and interpretability. However, most applications remain retrospective, limiting their use in scenarios requiring continuous monitoring of evolving patient trajectories.

A critical prerequisite for process mining is the construction of reliable event logs from heterogeneous data sources. In healthcare, this is particularly challenging due to fragmentation, missing values, and inconsistent timestamps. This transformation step, referred to as data lifting, is recognized as a central challenge~\cite{pm_augusto2019,pm_eventabstraction}. In this work, temporal reconstruction is treated as a task-aware transformation step aimed at producing representations suitable for predictive monitoring.

Predictive process monitoring estimates future states or outcomes of running process instances. Early work focused on time and performance prediction~\cite{pm_predictive2011}, while later approaches introduced machine learning and deep learning models for predicting activities and outcomes from trace prefixes~\cite{pm_tax2017,pm_evermann2017}. Prefix-based representations enable reasoning over partially observed traces, but most approaches emphasize predictive accuracy over reproducible data preparation and system integration.

Explainability is important in healthcare, where predictions must be understandable to domain experts. Prior work has explored feature-based and process-aware explanation techniques~\cite{pm_galanti2020,pm_rizzi2020}. Here, explainability is treated as a complementary direction supported by structured process-aware representations.

Overall, existing approaches remain largely retrospective, often decoupled from reproducible data pipelines, and rarely evaluate temporal reconstruction as part of the monitoring architecture. This paper addresses these limitations through an end-to-end, reproducible, process-aware pipeline integrating data lifting, prefix-based representations, and predictive monitoring.

\section{Proposed Pipeline}
\label{sec:pipeline}

The proposed framework extends traditional process mining pipelines by introducing a \textbf{process-aware and prefix-based monitoring architecture}. Unlike classical approaches focused on retrospective analysis of complete traces, the pipeline supports predictive monitoring on partially observed clinical pathways. The architecture transforms heterogeneous clinical data into temporally consistent event logs, from which prefix-based representations are derived to enable predictive modeling. The workflow is organized into four stages:
\begin{enumerate}
\item data integration and temporal validation;
\item event log construction and enrichment;
\item prefix-based process representation and feature extraction;
\item predictive monitoring.
\end{enumerate}

\begin{figure*}[t]
    \centering
    \includegraphics[width=0.99\linewidth]{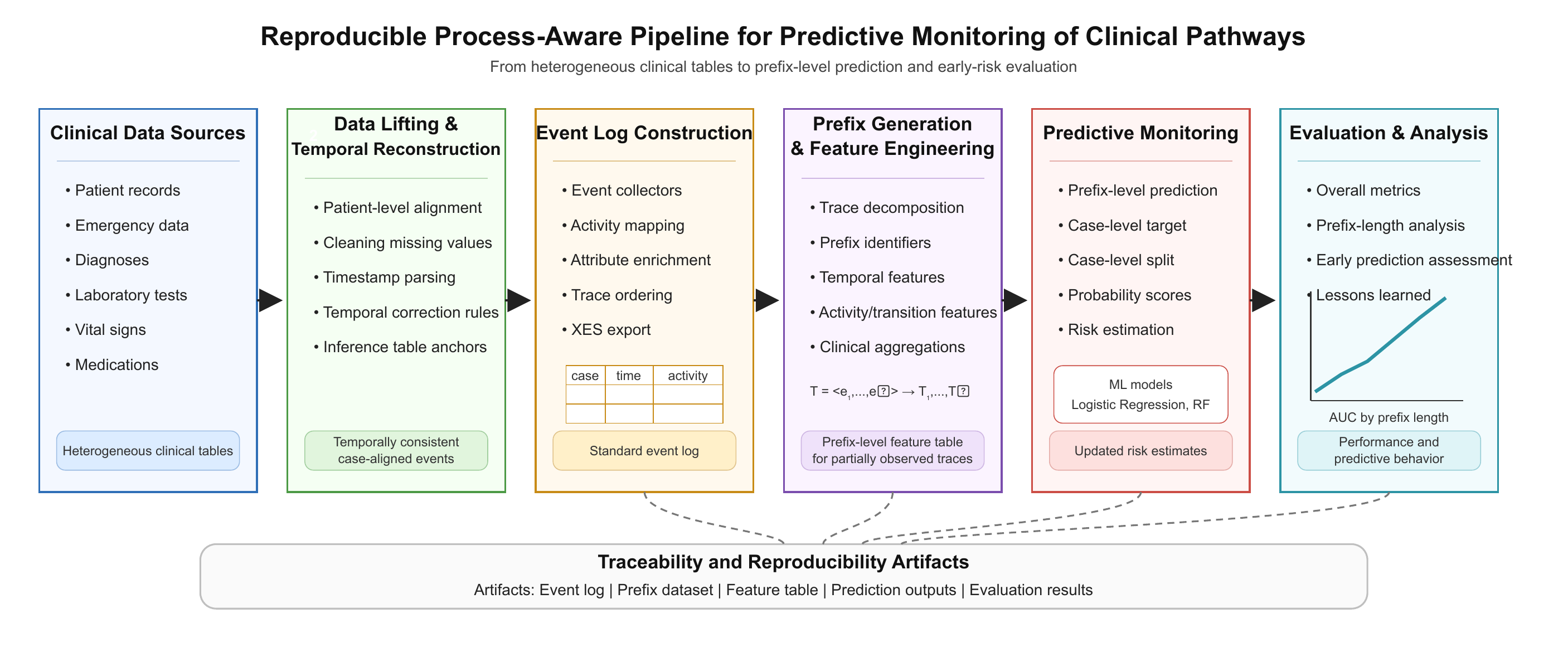}
    \caption{Proposed pipeline for predictive monitoring of clinical pathways.}
    \label{fig:pipeline}
\end{figure*}

Each stage is implemented as a modular component, ensuring reproducibility and traceability. Figure \ref{fig:pipeline} provides a refined view of the pipeline, where the four main stages described in the text are decomposed into finer-grained components for clarity.

\subsection{Data Integration and Temporal Validation}

Clinical data are integrated from multiple heterogeneous sources, including patient records, emergency diagnoses, hospital diagnoses, vital signs, medications, and laboratory tests. These data are stored in separate relational tables and are merged into a unified event-based representation through a deterministic data-lifting procedure implemented in \texttt{extract\_log.py}.

All source tables are aligned at the patient level using a unique patient identifier, which serves as the case identifier. To reduce redundancy, high-frequency tables such as vital signs and laboratory tests are optionally compressed by grouping records with identical patient identifiers and timestamps, retaining the first occurrence.

Data cleaning is applied to standardize missing and invalid values. In particular, zero values and empty strings are converted to missing values (NaN) to ensure consistency across heterogeneous sources.

Temporal validation and reconstruction are explicitly implemented through rule-based transformations. Timestamps are constructed by combining separate date and time attributes and converting them into a unified datetime format. Domain-informed correction rules are then applied to resolve inconsistencies. For example, triage timestamps that precede admission times within a limited time window are shifted forward, and inconsistencies between related events are corrected through conditional reordering.

To support timestamp reconstruction, an inference table is computed for each patient by aggregating candidate timestamps from multiple sources (e.g., admission, ICU discharge, vital signs, laboratory tests). This table defines temporal anchor points that are used to infer missing or partially specified timestamps while preserving ordering constraints. Events labeled as \textit{FIRST}, \textit{NOT\_FIRST}, or \textit{LAST} are adjusted relative to these anchors to ensure consistent sequencing.

This stage produces a temporally consistent, case-aligned representation of clinical events, preserving incomplete records while enforcing temporal coherence.

\subsection{Event Log Construction and Enrichment}

The validated data are transformed into an event log through a deterministic extraction procedure implemented in \texttt{extract\_log.py}. Each patient is modeled as a case, and clinical activities are mapped to events using modular event collectors. Each event collector specifies:
(i) the source table(s),
(ii) the timestamp attribute,
(iii) the activity name, and
(iv) the associated attributes.

Event extraction is implemented through reusable functions supporting single-table extraction, multi-table joins, and aggregation of repeated measurements. Each event is represented as a structured record containing:
\begin{itemize}
\item \textbf{case identifier};
\item \textbf{activity name};
\item \textbf{timestamp}.
\end{itemize}

Additional attributes are attached depending on the event type. For example, monitoring events include vital sign measurements, while admission events include diagnostic and administrative information. Attribute names are standardized during extraction.

Events are grouped by patient and sorted by timestamp to form traces. Trace-level attributes (e.g., demographics) are propagated to all events using dedicated extraction functions. The resulting event log is stored in tabular format and subsequently converted into the XES format using \texttt{xes\_from\_csv.py} and the \textit{pm4py} library. During conversion, attributes are mapped to XES-compliant fields (e.g., \texttt{concept:name}, \texttt{time:timestamp}), and both event-level and trace-level information are preserved.

\subsection{Prefix-Based Process Representation and Feature Extraction}

To enable predictive monitoring, each trace is decomposed into a set of prefixes using \texttt{build\_prefix\_dataset.py}. Given a trace
\[
T = \langle e_1, e_2, \ldots, e_n \rangle,
\]
prefixes are defined as
\[
T_k = \langle e_1, e_2, \ldots, e_k \rangle \quad \text{for } k = 1,\ldots,n.
\]

Prefix generation is implemented by iterating over each trace and incrementally constructing temporally ordered event sequences. Each prefix is assigned a unique identifier, its corresponding case identifier, and its prefix length.

Feature extraction is performed in \texttt{feature\_engineering.py}, where prefix-level feature vectors are computed by aggregating event-level information. The extracted features include:

\begin{itemize}
\item \textbf{Temporal features:} elapsed time from the first event, time since admission, and inter-event intervals;
\item \textbf{Activity features:} counts and frequencies of activities, and last observed activity;
\item \textbf{Transition features:} patterns of consecutive activities capturing short-term dynamics;
\item \textbf{Clinical features:} aggregated statistics (e.g., latest, minimum, maximum, average) of time-varying signals;
\item \textbf{Case attributes:} static demographic and contextual information.
\end{itemize}

The resulting dataset is a prefix-level tabular representation that enables the application of standard machine learning models while preserving process-aware and clinically relevant information.

\subsection{Predictive Monitoring}

The predictive component estimates the likelihood of a target outcome from partially observed process trajectories. Each complete trace is decomposed into prefixes, and each prefix is represented through the feature vectors described in the previous section.

Prefix-level feature vectors are used as input to classification models implemented in \texttt{train\_predictive\_models.py}. The prediction target is defined at the case level and propagated to all corresponding prefixes, enabling supervised learning on partially observed trajectories.

Predictions are computed as probability scores for the positive class. The framework supports standard evaluation metrics such as AUC, accuracy, precision, recall, and F1-score, and enables analysis of predictive performance as a function of prefix length.

\section{Case Study: COVID-19 Clinical Pathways}
\label{sec:casestudy}
The proposed framework is evaluated on COVID-19 clinical pathways derived from the COVID Data for Shared Learning (CDSL) dataset. The dataset includes heterogeneous clinical information such as emergency admissions, hospital stays, ICU transfers, laboratory tests, medications, and monitoring activities.

The data integration and preprocessing pipeline
produces a temporally consistent event log, which is
subsequently transformed into a prefix-level dataset.
The resulting dataset includes 4,479 patient cases and
46,804 prefixes. For predictive evaluation, data
are split at the case level into training and test sets,
with 896 patient cases used for testing, allowing the
evaluation of predictive monitoring under partially
observed conditions.

The prediction target is ICU admission, which represents
a minority outcome in the dataset. In the generated
prefix-level dataset, positive cases account for
approximately 12.6\% of the instances, resulting in a
moderately imbalanced classification problem. For this
reason, model evaluation is based not only on accuracy,
but also on metrics such as AUC, precision, recall, and
F1-score, which provide a more reliable assessment of
predictive performance under class imbalance.

\subsection{Prediction Task and Experimental Setup}

The predictive task consists in estimating the likelihood of ICU admission from partially observed patient trajectories. The prediction target is defined at the case level and propagated to all corresponding prefixes.

Each patient trace is decomposed into prefixes, and each prefix is represented using the feature vectors generated by the pipeline. The experimental setup simulates an online monitoring scenario, in which predictions are updated incrementally as new events become available.

To prevent information leakage, the dataset is split at the case level, ensuring that all prefixes derived from the same patient are assigned to either the training or the test set. Predictive models are trained on prefix-level feature vectors and evaluated on unseen patient cases.

Model performance is evaluated using standard classification metrics, including Area Under the ROC Curve (AUC), accuracy, precision, recall, and F1-score. The overall predictive results are reported in Table~\ref{tab:overall_metrics}.

\input{outputs/table_overall_metrics.tex}

The results show that Logistic Regression achieves the best overall performance, with an AUC of 0.906 and an F1-score of 0.835, outperforming the Random Forest model across most evaluation metrics. These results indicate that the prefix-based feature representation is sufficiently informative to support accurate prediction using relatively simple and interpretable models.

In addition to overall performance, predictive quality is analyzed as a function of prefix length to assess early prediction capability. The results are reported in Table~\ref{tab:prefix_metrics}.

\input{outputs/table_prefix_metrics.tex}

The prefix-level analysis highlights a progressive improvement in predictive performance as additional events become available. At early stages, performance is limited (AUC 0.642 at prefix length 1 and 0.658 at prefix length 5), reflecting the scarcity of available clinical information. As the pathway evolves, predictive performance increases steadily, reaching an AUC of 0.772 at prefix length 20 and exceeding 0.93 for longer prefixes.

These results confirm that the proposed prefix-based representation supports incremental risk estimation and enables the analysis of predictive behavior at different stages of the clinical pathway.

\subsection{Reproducibility and Experimental Artifacts}
The case study is designed to ensure full reproducibility of the experimental workflow. Each stage of the pipeline produces explicit intermediate artifacts that can be independently inspected and reused. In particular, the following artifacts are generated:
\begin{itemize}
\item \textbf{event log}: a temporally ordered representation of clinical events for each patient;
\item \textbf{prefix dataset}: a collection of partially observed trajectories, including case identifiers and prefix lengths;
\item \textbf{feature table}: a tabular representation of prefix-level features used for predictive modeling;
\item \textbf{prediction outputs}: model predictions, including probability scores and ground-truth labels for each prefix;
\item \textbf{evaluation results}: aggregated performance metrics computed overall and by prefix length.
\end{itemize}

All transformations are deterministic and implemented through explicit processing steps, ensuring that the entire workflow can be reproduced from the original clinical data. The explicit generation of intermediate artifacts allows tracing each prediction back to the corresponding prefix and underlying event sequence, providing full transparency of the data flow.

This artifact-based design enables not only reproducibility of the final results, but also fine-grained inspection of intermediate representations, supporting validation, debugging, and extension of the proposed pipeline in other clinical scenarios. The implementation will be made publicly available upon acceptance.

\section{Lessons Learned}

The experimental results provide several non-trivial findings that have implications for the design of process-aware digital health monitoring systems.

A first key observation is that predictive performance does not increase uniformly, but exhibits a delayed emergence of informative signals. Although early prefixes show limited discriminative power (AUC 0.642 at prefix length 1), substantial improvements are observed only after a sufficient accumulation of events, with performance exceeding AUC 0.77 around prefix length 20 and reaching 0.942 at prefix length 50. This suggests that clinically relevant predictive information is not concentrated at admission, but is progressively encoded in the sequence of events generated during the care process. As a consequence, predictive monitoring systems should be designed to continuously update risk estimates rather than relying on early-stage predictions alone.

A second important finding concerns the role of process structure in predictive performance. The consistent improvement across prefix lengths indicates that sequences of activities and their temporal ordering provide additional predictive power beyond isolated observations. This implies that process-aware representations are not merely an alternative encoding, but a necessary component for capturing the evolving clinical state. In particular, the results suggest that modeling transitions and temporal context is essential for achieving robust predictive performance in dynamic healthcare scenarios.

The results also highlight a structural trade-off between prediction earliness and reliability. Early-stage predictions are inherently affected by high uncertainty due to limited available information, while later predictions benefit from richer contextual evidence. However, the observed performance trajectory indicates that useful predictive signals already emerge at intermediate prefix lengths, suggesting that actionable predictions can be obtained before the completion of the clinical pathway. This has direct implications for the design of decision support systems, which must balance timeliness and confidence.

Another relevant finding concerns the impact of temporal reconstruction on predictive performance. The overall AUC of 0.906, achieved on a dataset with heterogeneous and partially inconsistent timestamps, indicates that accurate temporal alignment is a critical enabler for process-aware learning. This suggests that data preprocessing, and in particular temporal validation and reconstruction, should be considered an integral part of the predictive modeling pipeline rather than a preliminary step.

Finally, the generation of 46,804 prefixes from 896 patient cases demonstrates that prefix-based representations enable a substantial expansion of the training dataset while preserving process semantics. This increases the number of training instances without introducing artificial data, supporting both model training and fine-grained evaluation. At the same time, it enables detailed analysis of predictive performance across different stages of the clinical pathway, which would not be possible using only complete traces.

\section{Discussion}
\label{sec:discussion}

The results of this study highlight several implications for the design of predictive process monitoring systems in healthcare.

A first implication concerns the role of data transformation in predictive performance. The results indicate that the quality of temporal reconstruction and event log construction has a direct impact on model effectiveness. This suggests that improvements in predictive performance cannot be achieved only through more advanced learning algorithms, but also require careful design of the data-lifting pipeline. In this context, data preprocessing should not be considered a preliminary activity, but a central component of the predictive architecture.

A second implication relates to the use of process-aware representations. The observed performance improvements across prefix lengths confirm that modeling patient trajectories as evolving sequences provides additional predictive value compared to static representations. This supports the idea that process mining concepts can be effectively extended to predictive settings, where the objective is not only to analyze completed processes, but also to reason on running cases.

Another important observation concerns the relationship between representation quality and model complexity. Although Logistic Regression is simpler than Random Forest, it achieves slightly better overall performance. This suggests that the proposed prefix-level representation already captures a substantial part of the predictive information through process-aware feature engineering and temporal aggregation. In this setting, predictive signals appear to be strongly encoded in the constructed features, reducing the need for highly complex decision models.

This aspect is particularly relevant in healthcare environments, where simpler and more interpretable models can improve transparency and reproducibility. The use of interpretable models also facilitates traceability of the predictive process and supports clearer communication of results. More advanced temporal architectures, such as recurrent or transformer-based models, could potentially capture additional sequential dependencies that are simplified in the current tabular representation. However, the goal of this work is not to maximize predictive performance through increasingly complex architectures, but to evaluate the effectiveness of a reproducible and process-aware monitoring pipeline.

From a system design perspective, the results highlight the importance of incremental prediction mechanisms. Since predictive performance depends on the accumulation of evidence over time, monitoring systems should continuously update risk estimates rather than relying on single-point predictions. This has direct implications for real-world deployment, where clinical decisions must be made under uncertainty and refined as new information becomes available.

More broadly, the findings suggest a shift from model-centric approaches to pipeline-centric approaches. While much of the existing literature focuses mainly on algorithmic improvements, the results of this study show that the integration of data processing, temporal reconstruction, process-aware representations, and predictive modeling plays a decisive role in achieving reliable predictive performance. This perspective is particularly important in healthcare, where data heterogeneity and data quality issues are intrinsic characteristics of the domain.

Overall, these results support the idea that predictive monitoring in healthcare should be viewed as a continuous and process-aware activity, in which patient risk is progressively updated as clinical pathways evolve.

\subsection{Limitations}

The proposed approach involves several design choices motivated by the requirements of reproducibility, interpretability, and integration with process-aware representations. These choices also introduce some limitations that should be considered when interpreting the results.

First, the predictive models rely on tabular prefix-level features derived from aggregated event information. This representation enables the use of standard and interpretable machine learning models and provides a clear and deterministic mapping from event logs to predictive features, which is important for traceability in clinical applications. However, the current representation does not explicitly model fine-grained temporal dependencies that could potentially be captured by sequence-based approaches such as recurrent or transformer-based models. For the same reason, the study does not include a direct comparison with transformer-based predictive approaches, since the
focus is on evaluating the effectiveness of the proposed process-aware and reproducible pipeline rather than benchmarking advanced sequence architectures.

Second, the aggregation of event-level information into prefix-level descriptors summarizes temporal dynamics through statistical and process-aware features rather than explicit sequence modeling. While this design supports the integration of heterogeneous clinical signals within a unified feature space and enables scalable training and evaluation across large prefix datasets, some complex temporal patterns and long-range dependencies may not be fully represented.

Third, although the results demonstrate strong predictive performance, the study does not include a detailed clinical interpretation of the learned predictive patterns. The main objective of this work is the design of a reproducible and process-aware predictive monitoring pipeline rather than domain-specific clinical knowledge extraction. Additional validation with clinical experts would help assess the practical relevance, interpretability, and usability of the generated predictions in real-world healthcare settings.

Finally, the evaluation is conducted on COVID-19 clinical pathways derived from a single dataset. Although the dataset is large and heterogeneous, further experiments across different clinical domains and healthcare environments are necessary to assess the generalizability and robustness of the proposed approach.
\section{Conclusion}
\label{sec:conclusions}

This paper presented a reproducible and process-aware pipeline for predictive monitoring of clinical pathways based on partially observed patient trajectories. By integrating data lifting, temporal reconstruction, event log construction, prefix-based representations, and predictive modeling, the proposed framework enables the transition from retrospective process analysis to continuous, data-driven monitoring.

The experimental evaluation on COVID-19 clinical pathways shows that predictive performance improves as additional events become available, with the best-performing model achieving an AUC of 0.906 and an F1-score of 0.835. These results confirm that clinically relevant predictive signals emerge progressively along patient pathways and that process-aware representations are effective for modeling the temporal evolution of clinical states.

Beyond predictive performance, the results highlight a key insight: effective predictive monitoring in healthcare depends not only on the choice of learning algorithms, but also on the design of the data transformation pipeline. In particular, temporal alignment, consistent event representation, and prefix-based modeling play a central role in enabling reliable predictions from heterogeneous clinical data. This shifts part of the modeling effort from purely algorithmic optimization to the construction of reproducible and process-aware data pipelines.

Future work will focus on extending the proposed approach in three main directions: (i) deployment in streaming environments to support real-time monitoring and continuous model updating, (ii) evaluation across diverse clinical domains to assess robustness and generalizability, and (iii) integration of sequence-based models to capture complex temporal dependencies while preserving traceability and interpretability of the predictive process. 

\section*{\uppercase{Acknowledgements}}
We acknowledge financial support under the National Recovery and Resilience Plan (NRRP), M4C2I1.1, funded by the European Union – NextGenerationEU– Project Title aRtificial intElligence for Process Analytics (REPA) - Grant Assignment Decree No. 2022CJWPNA  (CUP Università degli Studi di Bari Aldo Moro H53C24000960006, CUP UNITELMA  I53C24002420006) by the Italian Ministry of University and Research (MUR).


\bibliographystyle{apalike}
{\small
\bibliography{example}}

\end{document}

%% file: outputs/table_overall_metrics.tex
\begin{table*}[h]
\caption{Overall predictive performance for ICU admission prediction.}
\label{tab:overall_metrics}
\centering
\begin{tabular*}{\textwidth}{|@{\extracolsep{\fill}}l|r|r|r|r|r|r|r|r|}
\hline
Model & Cases & Prefixes & Positive Rate & AUC & Accuracy & Precision & Recall & F1 \\
\hline
logreg & 896 & 46804 & 0.126 & 0.906 & 0.964 & 0.985 & 0.725 & 0.835 \\
\hline
rf & 896 & 46804 & 0.126 & 0.901 & 0.952 & 0.980 & 0.631 & 0.768 \\
\hline
\end{tabular*}
\end{table*}

%% file: outputs/table_prefix_metrics.tex

\begin{table*}[h]
\caption{Early prediction performance at selected prefix lengths using Logistic Regression.}
\label{tab:prefix_metrics}
\centering
\begin{tabular*}{\textwidth}{|@{\extracolsep{\fill}}r|r|r|r|r|r|}
\hline
Prefix length & Cases & AUC & Precision & Recall & F1 \\
\hline
1  & 896 & 0.642 & 0.714 & 0.152 & 0.250 \\
\hline
5  & 894 & 0.658 & 0.833 & 0.154 & 0.260 \\
\hline
10 & 879 & 0.716 & 0.947 & 0.286 & 0.439 \\
\hline
20 & 799 & 0.772 & 1.000 & 0.460 & 0.630 \\
\hline
30 & 658 & 0.826 & 0.970 & 0.561 & 0.711 \\
\hline
40 & 493 & 0.931 & 0.977 & 0.808 & 0.884 \\
\hline
50 & 349 & 0.942 & 0.975 & 0.867 & 0.918 \\
\hline
\end{tabular*}
\end{table*}

%% file: example.bib
@incollection{pm_aalst2016,
  title={Data science in action},
  author={Van Der Aalst, Wil},
  booktitle={Process mining: Data science in action},
  pages={3--23},
  year={2016},
  publisher={Springer}
}

@article{pm_rojas2016,
  author  = {Eric Rojas and Jorge Munoz-Gama and Marcos Sep{\'u}lveda and Daniel Capurro},
  title   = {Process Mining in Healthcare: A Literature Review},
  journal = {Journal of Biomedical Informatics},
  volume  = {61},
  pages   = {224--236},
  year    = {2016}
}

@article{pm_munoz2022,
  title={Process mining for healthcare: Characteristics and challenges},
  author={Munoz-Gama, Jorge and Martin, Niels and Fernandez-Llatas, Carlos and Johnson, Owen A and Sep{\'u}lveda, Marcos and Helm, Emmanuel and Galvez-Yanjari, Victor and Rojas, Eric and Martinez-Millana, Antonio and Aloini, Davide and others},
  journal={Journal of Biomedical Informatics},
  volume={127},
  pages={103994},
  year={2022},
  publisher={Elsevier}
}

@article{pm_augusto2019,
  author={Augusto, Adriano and Conforti, Raffaele and Dumas, Marlon and Rosa, Marcello La and Maggi, Fabrizio Maria and Marrella, Andrea and Mecella, Massimo and Soo, Allar},
  journal={IEEE Transactions on Knowledge and Data Engineering}, 
  title={Automated Discovery of Process Models from Event Logs: Review and Benchmark}, 
  year={2019},
  volume={31},
  number={4},
  pages={686-705},
  doi={10.1109/TKDE.2018.2841877}}

@article{pm_eventabstraction,
  title={Event abstraction in process mining: literature review and taxonomy},
  author={van Zelst, Sebastiaan J and Mannhardt, Felix and de Leoni, Massimiliano and Koschmider, Agnes},
  journal={Granular Computing},
  volume={6},
  number={3},
  pages={719--736},
  year={2021},
  publisher={Springer}
}

@article{pm_predictive2011,
  author  = {Wil M. P. van der Aalst and M. H. Schonenberg and Minseok Song},
  title   = {Time Prediction Based on Process Mining},
  journal = {Information Systems},
  volume  = {36},
  number  = {2},
  pages   = {450--475},
  year    = {2011}
}

@inproceedings{pm_tax2017,
  title={Predictive business process monitoring with LSTM neural networks},
  author={Tax, Niek and Verenich, Ilya and La Rosa, Marcello and Dumas, Marlon},
  booktitle={International conference on advanced information systems engineering},
  pages={477--492},
  year={2017},
  organization={Springer}
}

@article{pm_evermann2017,
title = {Predicting process behaviour using deep learning},
journal = {Decision Support Systems},
volume = {100},
pages = {129-140},
year = {2017},
note = {Smart Business Process Management},
issn = {0167-9236},
doi = {https://doi.org/10.1016/j.dss.2017.04.003},
url = {https://www.sciencedirect.com/science/article/pii/S0167923617300635},
author = {Joerg Evermann and Jana-Rebecca Rehse and Peter Fettke}
}

@inproceedings{pm_galanti2020,
  title={Explainable predictive process monitoring},
  author={Galanti, Riccardo and Coma-Puig, Bernat and de Leoni, Massimiliano and Carmona, Josep and Navarin, Nicol{\`o}},
  booktitle={2020 2nd International Conference on Process Mining (ICPM)},
  pages={1--8},
  year={2020},
  organization={IEEE}
}

@inproceedings{pm_rizzi2020,
  title={Explainability in predictive process monitoring: When understanding helps improving},
  author={Rizzi, Williams and Di Francescomarino, Chiara and Maggi, Fabrizio Maria},
  booktitle={International conference on business process management},
  pages={141--158},
  year={2020},
  organization={Springer}
}
